\documentclass{article}

\PassOptionsToPackage{numbers, compress}{natbib}

\usepackage[preprint]{neurips_2025}
\usepackage[preprint]{neurips_2025}

\usepackage[utf8]{inputenc} 
\usepackage[T1]{fontenc}    
\usepackage{hyperref}       
\usepackage{url}            
\usepackage{booktabs}       
\usepackage{amsfonts}       
\usepackage{nicefrac}       
\usepackage{microtype}      
\usepackage{xcolor}         
\usepackage{enumitem}
\usepackage{graphicx}
\usepackage{multirow}
\usepackage{tablefootnote}
\usepackage{colortbl}
\usepackage[most]{tcolorbox}
\definecolor{lightgray}{gray}{0.9}
\usepackage[hyphenbreaks]{breakurl}
\usepackage{pifont}
\usepackage{soul, color, xcolor}

\usepackage{xspace}

\definecolor{darkgreen}{RGB}{50,100,0}
\definecolor{darkred}{RGB}{200, 0, 0}
\newcommand{\cmark}{\textcolor{darkgreen}{\ding{52}}} %
\newcommand{\xmark}{\textcolor{darkred}{\ding{56}}}

\newcommand{\bmark}{\textcolor{darkred}{\ding{52}\rotatebox[origin=c]{-9.2}{\kern-0.7em\ding{55}}}}

\newcommand{\huggingface}{\raisebox{-1.5pt}{\includegraphics[height=1.05em]{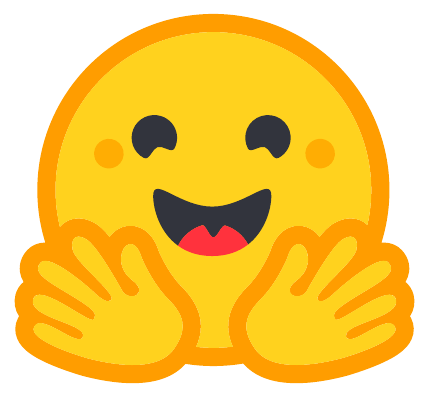}}\xspace}
\newcommand{\github}{\raisebox{-1.5pt}{\includegraphics[height=1.05em]{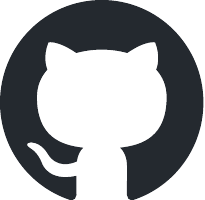}}\xspace}
\newcommand{\demo}{\raisebox{-1.5pt}{\includegraphics[height=1.05em]{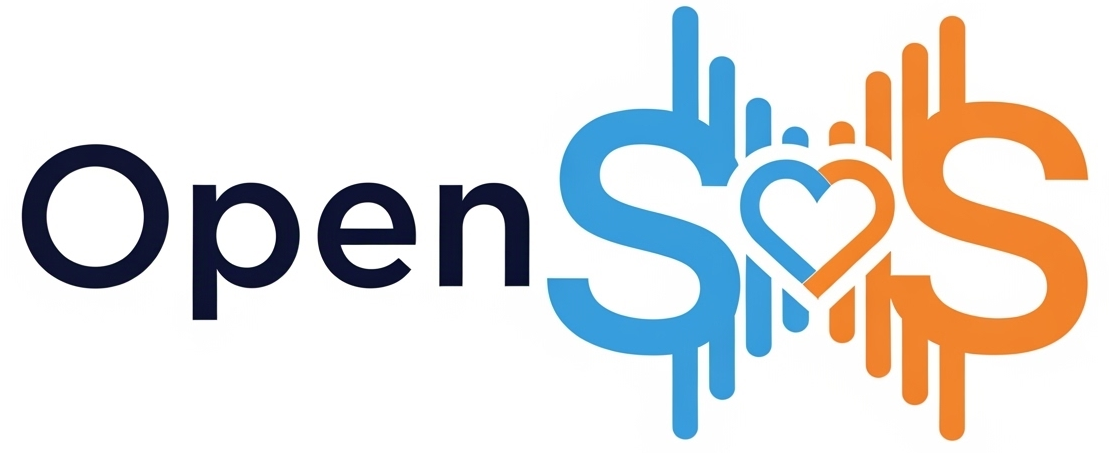}}\xspace}
\newcommand{\Openlogo}{\raisebox{-.5em}{\rlap{{\textcolor{white}{OpenS2S}}}\includegraphics[height=1.8em]{picture/opens2s_logo.png}}\xspace}

\title{\Openlogo{}: Advancing Fully Open-Source End-to-End Empathetic Large Speech Language Model}

\author{
Chen Wang\textsuperscript{1,2}, Tianyu Peng\textsuperscript{1,2,3}, Wen Yang\textsuperscript{1,2}, Yinan Bai\textsuperscript{1,2}, \\
\bf{Guangfu Wang}\textsuperscript{4}, Jun Lin\textsuperscript{4}, Lanpeng Jia\textsuperscript{4}, \\
\bf{Lingxiang Wu}\textsuperscript{1,3}, Jinqiao Wang\textsuperscript{1,2,3}, Chengqing Zong\textsuperscript{1,2}, Jiajun Zhang\textsuperscript{1,2,3} \thanks{\ \ Corresponding author} \\
\\
\textsuperscript{1}~Institute of Automation, Chinese Academy of Sciences \\
\textsuperscript{2}~School of Artificial Intelligence, University of Chinese Academy of Sciences \\
\textsuperscript{3}~Wuhan AI Research~~\textsuperscript{4}~GWM AI Lab \\
\texttt{wangchen2020@ia.ac.cn}~~\texttt{jjzhang@nlpr.ia.ac.cn}
}

\begin{document}

\maketitle

\begin{abstract}
Empathetic interaction is a cornerstone of human-machine communication, due to the need for understanding speech enriched with paralinguistic cues and generating emotional and expressive responses.
However, the most powerful empathetic LSLMs are increasingly closed off, leaving the crucial details about the architecture, data and development opaque to researchers.
Given the critical need for transparent research into the LSLMs and empathetic behavior, 
we present \texttt{OpenS2S}, a fully open-source, transparent and end-to-end LSLM designed to enable empathetic speech interactions. Based on our empathetic speech-to-text model BLSP-Emo~\cite{wang2024blsp}, \texttt{OpenS2S} further employs a streaming interleaved decoding architecture to achieve low-latency speech generation. To facilitate end-to-end training, \texttt{OpenS2S} incorporates an automated data construction pipeline that synthesizes diverse, high-quality empathetic speech dialogues at low cost. By leveraging large language models to generate empathetic content and controllable text-to-speech systems to introduce speaker and emotional variation, we construct a scalable training corpus with rich paralinguistic diversity and minimal human supervision.
We release the fully open-source \texttt{OpenS2S} model, including the dataset, model weights, pre-training and fine-tuning codes, to empower the broader research community and accelerate innovation in empathetic speech systems.
\begin{center}
\begin{tabular}{rcp{10cm}}
\multirow{1}{*}{\demo} & \textbf{Demo} & \href{https://casia-lm.github.io/OpenS2S}{\path{https://casia-lm.github.io/OpenS2S}}\\[0.2em]
\multirow{1}{*}{\github} & \textbf{Code} & \href{https://github.com/CASIA-LM/OpenS2S}{\path{https://github.com/CASIA-LM/OpenS2S}}\\[0.2em]
\multirow{3}{*}{\huggingface} & \textbf{\texttt{OpenS2S}} & \href{https://huggingface.co/CASIA-LM/OpenS2S}{\path{https://huggingface.co/CASIA-LM/OpenS2S}}\\[0.2em]
& \textbf{\texttt{OpenS2S\_V1.5}}\tablefootnote{\ \ [October 2025] We have released \texttt{OpenS2S\_V1.5}, which demonstrates significant improvements in performance across multiple metrics and practical experiences compared to the original version.} & \href{https://huggingface.co/CASIA-LM/OpenS2S_V1.5}{\path{https://huggingface.co/CASIA-LM/OpenS2S_V1.5}}\\[0.2em]
& \textbf{Data} & \href{https://huggingface.co/datasets/CASIA-LM/OpenS2S_Datasets}{\path{https://huggingface.co/datasets/CASIA-LM/OpenS2S_Datasets}}\\[0.2em]
\end{tabular}
\end{center}
\end{abstract}

\begin{table}[htbp]
\centering
\footnotesize
\setlength{\tabcolsep}{1mm}
\renewcommand\arraystretch{1.2}
\caption{\label{tab:intro_coparison_openness} The degree of openness of Open LSLMs.}
\begin{tabular}{l|cccccc}
    \toprule[1.2pt] 
    \textbf{Name} & LLaMA-Omni2 & Qwen2-Audio & GLM-4-Voice & Kimi-Audio & \texttt{OpenS2S} \\ 
    \midrule[0.8pt]
    Training Data & \xmark & \xmark  & \xmark & \xmark &\cmark \\
    Pre-training Code & \xmark & \xmark  & \xmark & \xmark &\cmark \\
    Fine-tuning Code & \xmark & \xmark  & \xmark & \cmark &\cmark \\
    Model & \cmark & \cmark  & \cmark & \cmark &\cmark \\
    Empathetic & \xmark & \xmark  & \cmark & \cmark & \cmark \\
    \bottomrule[1.2pt]
\end{tabular}
\end{table}

\section{Introduction}

Empathy is a fundamental pillar of human interactions, fostering everything from prosocial behavior to deeper connections~\cite{morelli2015emerging}.
Modeling and understanding empathy is a complex task for artificial intelligence, yet its integration is crucial for fostering more natural and effective human-machine communication~\cite{rashkin2018towards}.
In the realm of Large Speech Language Models (LSLMs), this challenge is particularly pronounced. 
Speech inherently conveys a wealth of rich paralinguistic information, including intonation, rhythm, volume variations, and cues related to speaker attributes like age and gender.
This intricate paralinguistic content makes speech communication highly sensitive, rendering flat or unnuanced responses from automated systems unacceptable. 
Consequently, developing empathetic speech systems is vital for creating more natural and human-centered artificial intelligence.

While recent advancements in LSLMs have significantly enhanced audio processing and enabled robust semantic-based instruction following in conversations~\cite{kannan2019large,radford2023robust,kheddar2024automatic,hao2023boosting,li2023styletts,barrault2023seamless,huang2023speech}, most existing models tend to overlook critical paralinguistic information in speech, thereby fundamentally limiting their native empathetic interaction capabilities.
Although some LSLMs~\cite{defossez2024moshi, zeng2024glm, ding2025kimi} demonstrate strong empathetic performance, they typically necessitate extensive pre-training on millions of hours of high-quality speech data. This reliance on vast datasets incurs substantial annotation, computation, and training costs, setting up a significant barrier to their broader adoption and development.
Furthermore, many of the most advanced models, particularly commercial models like GPT-4o~\cite{hurst2024gpt} and Gemini are fully proprietary and closed-source. This lack of transparency makes it challenging to analyze their internal mechanisms, reproduce their empathetic behaviors, or build upon their architectures for further scientific research and development. To scientifically study the empathetic behaviors in LSLMs, including potential biases, cultural variations, and their ethical implications, we believe that access to powerful, fully open empathetic LSLMs is critical to the advancement of this field.

To address the aforementioned limitations, we propose \texttt{OpenS2S}, a fully open-source, end-to-end LSLM. \texttt{OpenS2S} not only exhibits competitive foundational speech capabilities but also features an efficient streaming architecture based on interleaved decoding. Crucially, in contrast to existing models that achieve empathetic capabilities through resource-intensive pre-training, \texttt{OpenS2S} attains comparable empathetic interaction performance with significantly lower training data and computational costs. Moreover, the empathetic support provided by \texttt{OpenS2S} transcends mere paralinguistic cues, extending deeply into the semantic content of the dialogue. 


Overall, our main contributions are as follows:
\begin{enumerate}
\item \textbf{Model Construction and Training:} We build an efficient speech-to-speech empathetic model based on an advanced framework and conduct extensive training using high-quality data. This model can provide a more convenient and natural way for humans to interact with artificial intelligence.
\item \textbf{Automatic Empathetic Speech Instruction Dataset Construction:} We propose a data augmentation method for empathetic speech dialogue by combining the strengths of large language models (LLMs) and text-to-speech (TTS) models. LLMs are used to generate diverse user queries and empathetic responses, while voice cloning ensures input speaker diversity. InstructTTS further enables controllable emotional expression in speech responses, facilitating the construction of rich, high-quality training data with minimal human annotation.
\item \textbf{Fully Open-Source Release:} 
To foster collaborative research and accelerate innovation in empathetic LSLMs, we release all the resources, including the model weights, all codes for constructing datasets, pre-training, fine-tuning and evaluation, and the synthetic datasets, providing fully transparency and reproducibility for the community.
\end{enumerate}

\section{Method}

\subsection{Architecture}

\begin{figure}[t]
    \centering
    \includegraphics[width=12cm]{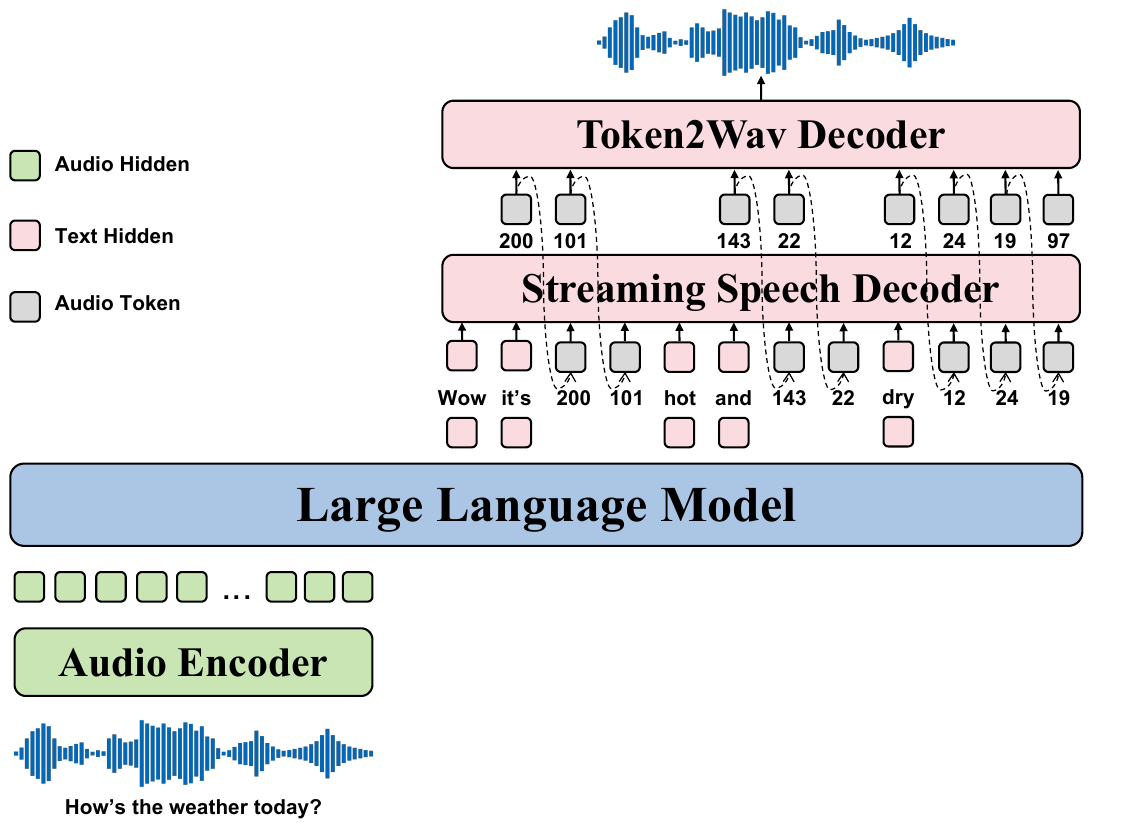}
    \caption{Architecture of our proposed \texttt{OpenS2S}.}
    \label{fig:Architecture_structure}
\end{figure}

The \texttt{OpenS2S} model architecture is shown in Figure~\ref{fig:Architecture_structure}, comprising four components: an audio encoder, an instruction-following LLM, a streaming speech decoder, and a token2wav decoder. Next, we will describe how to understand continuous speech input and ultimately generate an empathetic speech response.

\paragraph{Audio Encoder}

The Audio Encoder is responsible for transforming this raw audio signal into a more manageable and meaningful representation. To achieve this we use the encoder of Qwen2-Audio~\cite{chu2024qwen2} to extract features from the audio waveform due to its powerful ability to encode semantic content and paralinguistic information. These features generated by the audio encoder are encoded at a frequency of 25Hz. To further reduce the sequence length, the encoded representations are fed into a speech adapter, which comprises a downsampling module and a feed-forward network. The downsampling module, consisting of two CNN layers, is designed to compress the sequence length by a factor of 4. Finally, the features output by the speech adapter yield continuous encoded representations at 6.25Hz.

\paragraph{Instruction-Following LLM} 

The audio embeddings and text embeddings are concatenated to form interleaved input sequences for the large language model. We select Qwen3-8B-Instruct~\cite{yang2025qwen3} as the LLM, leveraging its robust text processing capabilities.

\paragraph{Streaming Speech Decoder} 

To enable streaming speech generation, we adopt a framework inspired by Minmo~\cite{chen2025minmo} and LLaMA-Omni2~\cite{fang2025llama}. The speech response is first converted into discrete tokens using a supervised semantic speech tokenizer. Then, an autoregressive text-to-speech language model is used to generate speech tokens conditioned on the hidden states of the LLM, enabling real-time generation.

The speech tokenizer is implemented by inserting a quantization module into the encoder of Whisper-large-v3~\cite{radford2023robust}, ultimately producing a token sequence at a resolution of 12.5 tokens per second with a vocabulary size of 16,384. We leverage the pretrained speech tokenizer from GLM-4-Voice~\cite{zeng2024glm}.

Once the speech response is tokenized, a decoder-only Transformer models the conditional generation from LLM hidden states to speech tokens. This decoder is initialized from Qwen3-1.8B, with its vocabulary extended to include the 16,384 speech tokens. The input to the streaming speech decoder consists of the final hidden states from the LLM, which are first projected via a linear layer to match the embedding dimension of the speech decoder.

To achieve streaming generation, we interleave the LLM hidden states and generated speech tokens in a predefined ratio: for every $M$ hidden states consumed, $N$ speech tokens are generated (in our implementation, $M = 4$ and $N = 8$). After all hidden states are consumed, the model continues to autoregressively generate the remaining speech tokens until the response is complete. During training, the cross-entropy loss is computed only on the generated speech tokens.

\paragraph{Token2Wav Decoder}

The speech tokens generated by the streaming speech decoder are subsequently converted into the final speech waveform by the token2wav decoder. This module comprises two key components: a chunk-aware causal flow matching model, which incrementally synthesizes every M speech tokens into mel-spectrograms in a streaming fashion, and a HiFi-GAN vocoder, which converts the mel-spectrograms into the final waveform. Both the flow matching model and the vocoder are adopted from the pretrained components in GLM-4-Voice~\cite{zeng2024glm}.

\subsection{Training Strategy}

The training of \texttt{OpenS2S} consists of three stages: speech understanding pre-training, speech generation pre-training, and empathy speech instruction fine-tuning. In the first two pre-training stages, we utilize open-source ASR and TTS datasets for pre-training to endow the model with robust speech understanding and generation capabilities. In the instruction fine-tuning stage, we construct an empathy speech instruction dataset for fine-tuning, enabling the model to understand the semantic content and paralinguistic cues in speech, and finally generate empathic speech responses.

\begin{figure}[t]
    \centering
    \includegraphics[width=14cm]{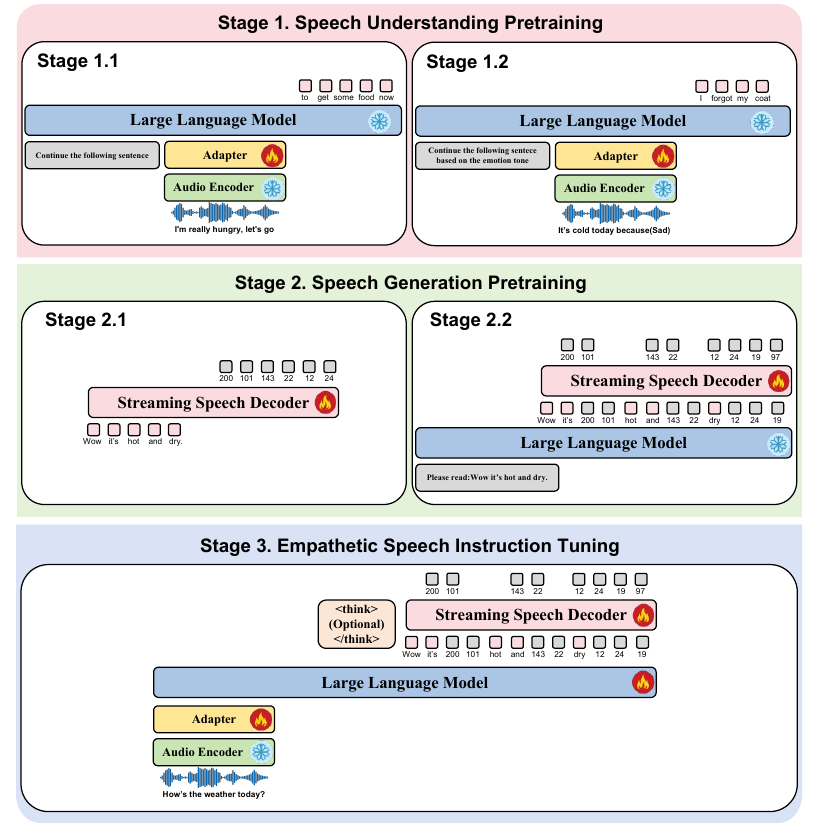}
    \caption{The training process of \texttt{OpenS2S}.}
    \label{fig:Training_Procedure}
\end{figure}

\paragraph{Speech Understanding Pretraining (Stage 1 in Figure~\ref{fig:Training_Procedure})}

To equip the model with robust speech understanding capabilities, we perform pretraining on large-scale speech-text paired corpora, following the training paradigm of BLSP-Emo~\cite{wang2024blsp}. This stage is divided into two phases: semantic alignment and emotional alignment.

In the semantic alignment phase, we adopt the concept of behavioral alignment, requiring the LLM to produce identical continuations when given either speech or its corresponding transcript as input. Specifically, we first prompt the LLM to generate continuations based on text transcripts from an ASR dataset. During training, the model is required to produce the same continuation when conditioned on continuous speech representations, thus aligning the model’s behavior across modalities.

In the emotional alignment phase, we leverage an SER dataset where each transcript is annotated with an emotion label. An LLM is prompted to generate emotion-aware continuations based on the transcript and the reference emotion. We then adapt a speech-language model to generate similar continuations directly from the speech input. This step encourages the model to comprehend and reflect both linguistic semantics and paralinguistic emotional cues, producing text responses aligned with those generated by the LLM given identical content and emotion labels.

Throughout this pretraining process, the parameters of the audio encoder and the LLM remain frozen. Only the speech adapter is fine-tuned to facilitate modality bridging.

\paragraph{Speech Generation Pretraining (Stage 2 in Figure~\ref{fig:Training_Procedure})}

As the streaming speech decoder is initialized from Qwen3-1.8B, which is originally designed for text generation, we first perform offline TTS pretraining to enable it to generate discrete speech tokens. Specifically, we expand the decoder’s vocabulary to include 16,384 speech tokens. During this phase, the input text is embedded using word embeddings as a prefix, and the target is the sequence of speech tokens extracted from reference audio. This allows the decoder to learn a basic mapping from text tokens to speech tokens, serving as a foundation for downstream speech synthesis.

In the second stage, we further train the speech decoder to integrate with the LLM for streaming generation. Unlike the previous step, the input text is no longer embedded directly into the speech decoder. Instead, it is first processed by the LLM using a structured prompt. The final-layer hidden states corresponding to the response portion are then extracted and interleaved with speech tokens during training. At this stage, the LLM’s parameters are kept frozen, while the linear projection layer and the speech decoder are fine-tuned. This training strategy not only bridges the LLM and the speech decoder but also adapts the offline decoder into a streaming-capable model that supports interleaved token generation.

\paragraph{Empathetic Speech Instruction Tuning (Stage 3 in Figure~\ref{fig:Training_Procedure})}

Following the pretraining stages, the model demonstrates general speech understanding capabilities, enabling it to generate empathetic text responses conditioned on both semantic content and emotional cues in speech. However, its speech generation ability remains limited: the model is only able to produce meaningful speech tokens when explicitly prompted with speech synthesis tasks. When handling general-purpose text instructions or directly responding to speech instructions, the speech decoder often fails to generate coherent or meaningful speech outputs.

We attribute this limitation to overfitting during the TTS-based pretraining stage. Specifically, the model learns to rely on a narrow representation subspace defined by the TTS task, resulting in poor generalizability to broader instruction-following scenarios. To address this issue, we introduce an additional instruction tuning stage aimed at enabling robust and flexible speech generation across diverse task types.

In this stage, the speech encoder remains frozen, while all other components, including audio adapter, the LLM, linear projection layer, and speech decoder, are fully fine-tuned. We observe that relying solely on speech-to-speech instruction data is insufficient to generalize speech generation to textual instructions. Therefore, we further incorporate text-to-speech instruction data, allowing the model to handle both speech and text inputs seamlessly. The construction process of these instruction datasets is detailed in Section~\ref{sec:sft_data}.

\section{Data Collection}

\subsection{Pre-training}

\paragraph{Speech Understanding}

For the semantic alignment stage, we utilize publicly available ASR datasets, and for the emotion alignment stage, we employ standard SER datasets. The ASR datasets include LibriSpeech~\cite{panayotov2015librispeech}, CommonVoice 13.0~\cite{ardila2019common}, and the GigaSpeech~\cite{chen2021gigaspeech} M subset, comprising approximately 1.9 million English (speech, text) pairs. A comparable number of Chinese ASR samples are randomly drawn from WeNetSpeech~\cite{zhang2022wenetspeech}. The SER datasets consist of IEMOCAP~\cite{busso2008iemocap}, MELD~\cite{poria2018meld}, CMU-MOSEI~\cite{zadeh2018multimodal}, MEAD~\cite{wang2020mead}, and ESD~\cite{zhou2022emotional}, collectively covering around 70k utterances in both English and Chinese.

\paragraph{Speech Generation} 

In the first stage of pretraining, aimed at expanding the vocabulary and enhancing the speech decoder’s capacity to generate speech tokens, we use 5k hours of English and 5k hours of Chinese speech randomly sampled from the Emilia~\cite{he2024emilia} dataset. For connecting the large language model with the speech decoder and adapting it into an interleaved streaming generation decoder, we further sample 1k hours each of English and Chinese data from the first-stage dataset for training.

\subsection{Supervised Fine-tuning}
\label{sec:sft_data}

Existing open-source speech instruction datasets commonly face three key challenges:
\begin{itemize}
    \item Limited speaker diversity, which undermines model robustness to varied speech inputs. For example, datasets such as InstructS2S-200K~\cite{fang2024llama} and E-Chat~\cite{xue2024chat} rely on TTS systems to synthesize speech from text, resulting in limited variability in speaker characteristics.
    \item Neglect of paralinguistic information, with an exclusive focus on semantic content. Although VoiceAssistant-400K~\cite{xie2024mini} introduces speaker diversity via voice cloning, it fails to capture critical paralinguistic cues such as emotion and speaking style.
    \item Insufficient label granularity, as exemplified by SD-Eval~\cite{ao2024sd}, which is divided into subsets that annotate different paralinguistic attributesfeatures in isolation—such as emotion or gender—in isolation. Nowithout any single subset providesoffering joint annotations across multiple paralinguistic dimensions.
\end{itemize}

To address these limitations, we propose a fully automated framework for constructing an empathetic speech instruction dataset. This framework systematically enhances speech diversity and representativeness across dimensions such as emotion, age, and gender through the integration of heterogeneous data sources. The full pipeline is illustrated in Figure~\ref{fig:data_structure} and comprises the following three stages:

\begin{figure}[tp]
    \centering
    \includegraphics[width=12cm]{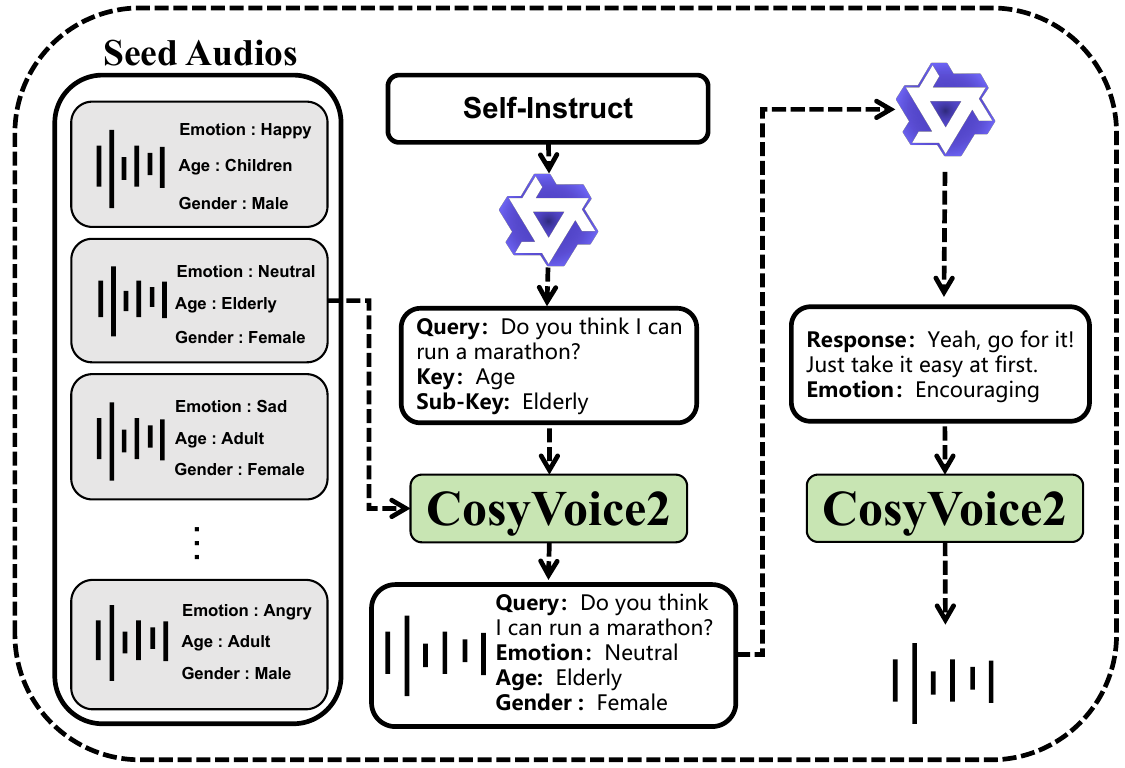}
    \caption{Automatic workflow for constructing empathetic speech instruction dataset.}
    \label{fig:data_structure}
\end{figure}

\paragraph{Collection and Manual Annotation of Seed Audio}

We begin by selecting seed audio samples from several publicly available speech emotion recognition datasets. These datasets cover a wide range of emotions and speaker demographics, including children, adults, and the elderly, ensuring strong representativeness and diversity. Each selected sample is manually annotated with its transcribed text, speaker gender, age, and emotional label. This results in 1,000 English and 1,000 Chinese seed audio samples with rich multi-dimensional annotations.

\paragraph{Generation of Speech Instructions} 

A straightforward solution to generating speech instructions is to convert existing text instruction datasets into speech. However, such datasets present two challenges: (1) tasks involving math or programming are often unsuitable for conversational speech scenarios; and (2) most instruction data neglect paralinguistic factors. Inspired by the Self-Instruct paradigm, we leverage Qwen3-32B-Instruct to automatically generate task instructions that are sensitive to paralinguistic cues. For example, the model might generate the instruction "Do you think I can run a marathon?" and tag it as age-sensitive, suggesting it be read in an elderly voice. We then randomly select a seed audio whose emotion label is "elderly" as an audio prompt and use CosyVoice2~\cite{du2024cosyvoice} for voice cloning, preserving the emotion and gender attributes of the selected seed. This process yields 50k English and 50k Chinese speech instructions with diverse paralinguistic characteristics.

\paragraph{Generation of Speech Response} 

For each speech instruction, we annotate its transcribed text, emotion, age, and gender based on the matched seed audio. Both the semantic content and paralinguistic labels are input into Qwen3-32B-Instruct with "thinking mode" enabled. Inspired by LLaMA-Omni, we prompt the model to generate concise, dialogue-appropriate, and empathetic text responses. We then prompt the same model to infer the appropriate emotional tone for delivering the response, conditioned on the original instruction, the response content, and the paralinguistic features of the input speech. Finally, we use a consistent reference voice as a prompt and control CosyVoice2 via instruction prompts to synthesize emotionally expressive speech responses.

\begin{figure}[t]
    \centering
    \begin{minipage}[t]{0.24\linewidth}
        \centering
        \includegraphics[width=\linewidth]{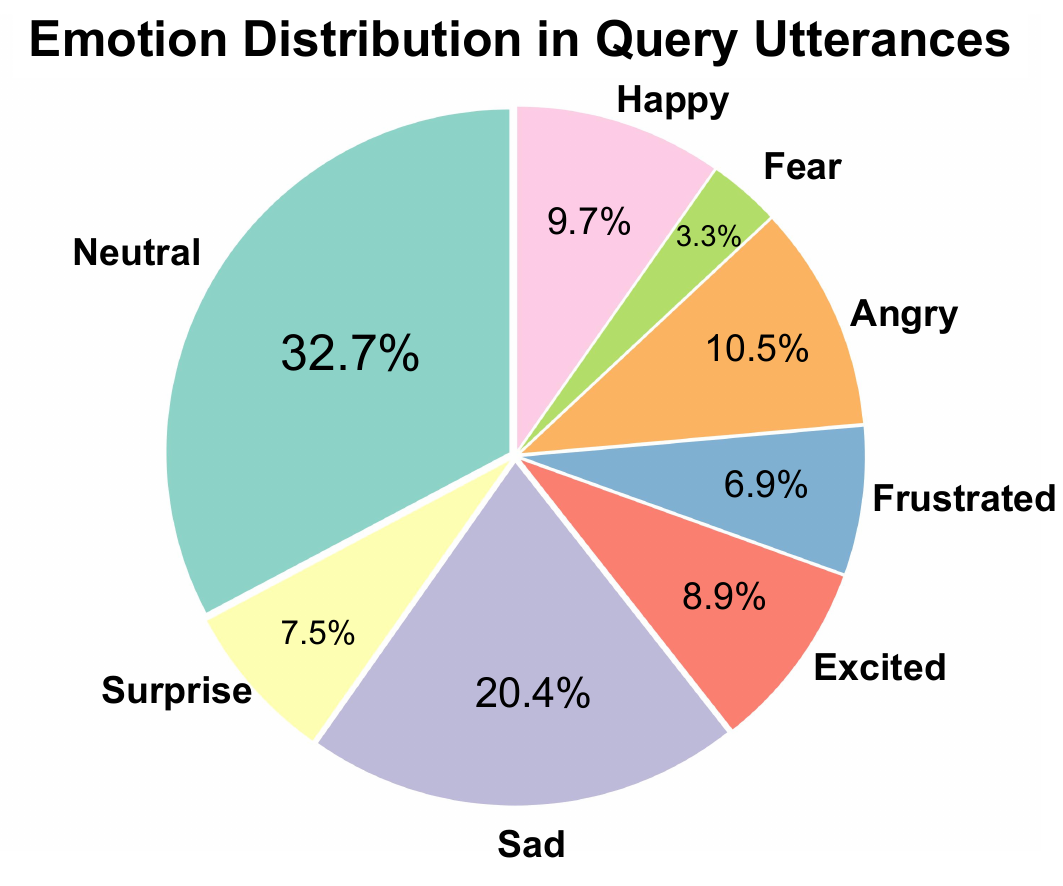}
    \end{minipage}
    \begin{minipage}[t]{0.24\linewidth}
        \centering
        \includegraphics[width=\linewidth]{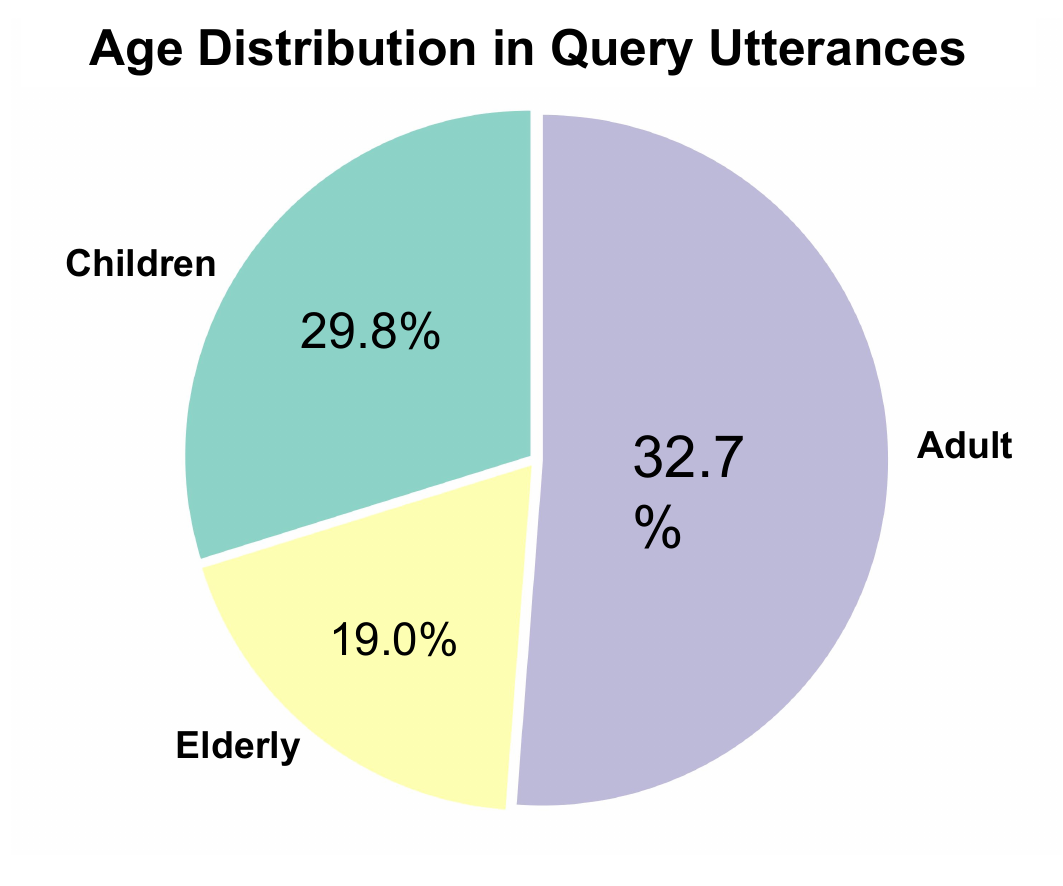}
    \end{minipage}
    \begin{minipage}[t]{0.24\linewidth}
        \centering
        \includegraphics[width=\linewidth]{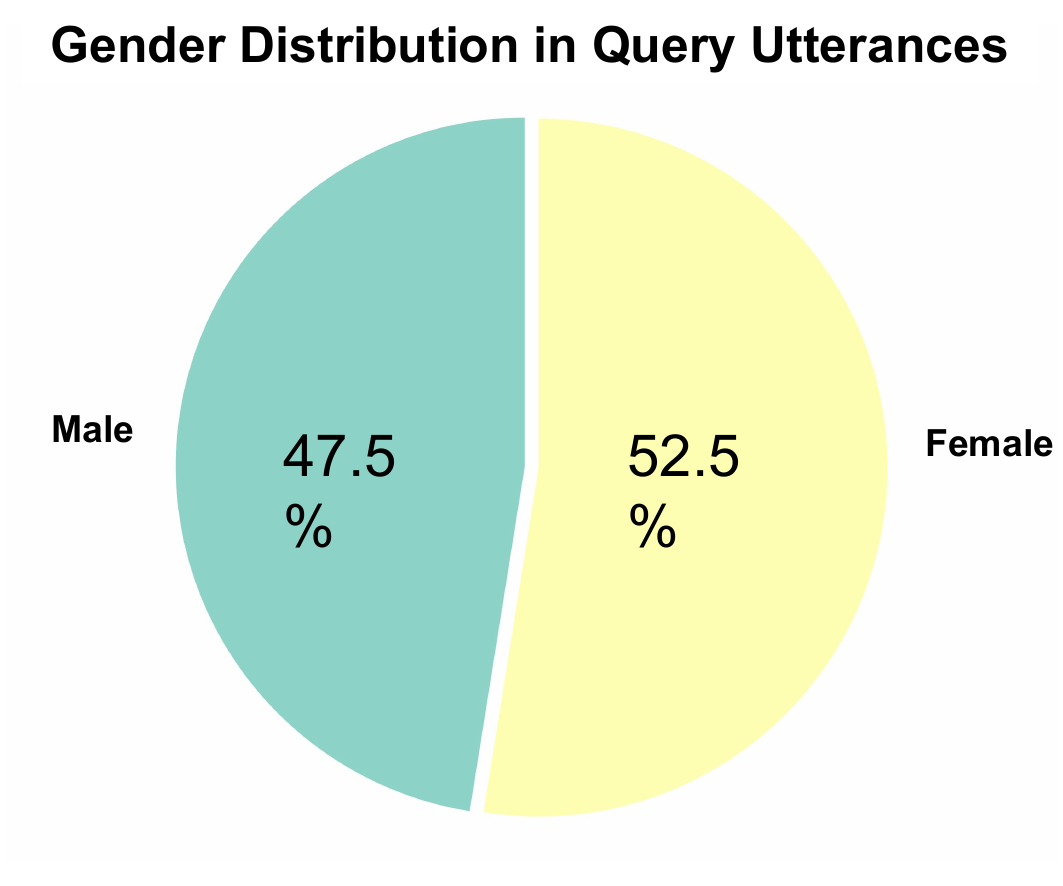}
    \end{minipage}
    \begin{minipage}[t]{0.24\linewidth}
        \centering
        \includegraphics[width=\linewidth]{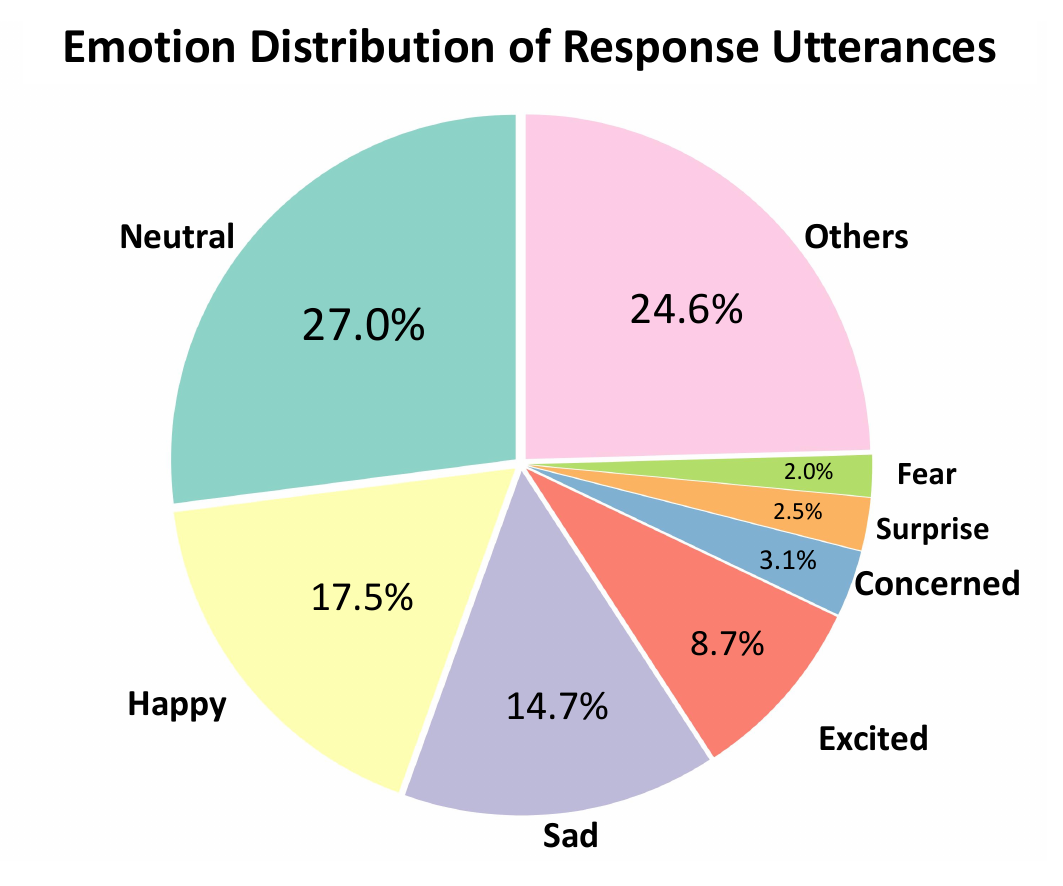}
    \end{minipage}
    \vspace{2mm}
    \caption{Distributions of emotion, age, and gender in query utterances, and emotion distribution in response utterances.}
    \label{fig:all_stats}
\end{figure}

Through this systematic construction process, we obtain an empathetic speech instruction dataset characterized by multi-dimensional emotion annotation, expressive emotional delivery, and diverse speaker profiles. The statistics of the constructed data are presented in Figure~\ref{fig:all_stats}. In total, we construct 50k English and 50k Chinese speech-to-speech empathetic samples. The input speech includes three types of paralinguistic information tags (i.e. emotion, gender, age), and the output speech is fixed as a young female voice responding with different emotions. [\textcolor{red}{\textbf{UPDATE (October 2025)}} In the V1.5 version, we expand the training data using the same approach, introducing approximately 200K English and 200K Chinese speech-to-speech empathetic samples.]

To retain the model’s general instruction-following capabilities beyond empathetic dialogue, we further incorporate general-purpose data. Specifically, we extract 50k English instructions from Instruct200K~\cite{fang2024llama}, translate them into Chinese using Qwen3-32B-Instruct, and select seed audio with neutral emotional labels to convert these instructions into speech. Applying the same speech response generation process as above, we obtain an additional 100k bilingual speech-to-speech instruction pairs.[\textcolor{red}{\textbf{UPDATE (October 2025)}} In the V1.5 version, we incorporate all data from Instruct200K, resulting in 400K bilingual speech-to-speech instruction pairs.]

Lastly, we observe that training solely on speech-to-speech data can impair the model’s ability to respond to text inputs: while the language model remains capable of generating reasonable text, the speech decoder fails to produce valid speech tokens when conditioned on text instruction. To mitigate this, we extract text-to-speech instruction samples from the general speech-to-speech dataset, ensuring the model can jointly handle both text and speech inputs during inference.

\section{Evaluation}

\subsection{Speech-to-Text Chat}


We evaluate the ability of \texttt{OpenS2S} to engage in speech-to-text conversations based on audio input using two benchmarks:

\begin{itemize}
    \item \textbf{VoiceBench}~\cite{chen2024voicebench} is a benchmark designed for the multi-faceted evaluation of LLM-based voice assistants. To assess the crucial capability of instruction-following, we utilize the alpacaeval, commoneval, wildvoice, and ifeval subsets. These were specifically chosen to test the model's ability to comprehend and accurately execute diverse spoken commands.
    \item \textbf{URO-Bench}~\cite{yan2025uro} is an end-to-end benchmark for spoken dialogue models that assesses understanding, reasoning, and oral conversation skills, including paralinguistic cues. To evaluate the model's capacity for empathy, we employ its UnderEmotion-en and UnderEmotion-zh subsets. These are designed to measure the model's ability to perceive the user's emotional state and generate affectively appropriate responses in both English and Chinese.
\end{itemize}

These two benchmarks evaluate from multiple perspectives to ensure comprehensive assessment: VoiceBench assesses the model’s instruction-following capability, while URO-Bench evaluates its comprehension and response to paralinguistic emotional cues. We also evaluate several mainstream open-source speech large models with comparable parameter sizes to \texttt{OpenS2S} for comparison. The results are presented in Table~\ref{tab_speech2text}.

\begin{table}[]
    \centering
    \caption{Performance of \texttt{OpenS2S} and baseline models on the benchmarks of speech-to-text chat.}
    \label{tab_speech2text}
    \begin{tabular}{l|cccc|cc}
        \toprule[1.2pt] 
        \multirow{2}{*}{\textbf{Model}} & \multicolumn{4}{c|}{\textbf{VoiceBench}} & \multicolumn{2}{c}{\textbf{URO-Bench}} \\
        \cline{2-7}
        & alpaca & common & ifeval & wildvoice & underemo-en & underemo-zh \\
        \midrule[0.8pt]
        Qwen-2-Audio~\cite{chu2024qwen2} & 3.74 & 3.43 & 26.33 & 3.01 & 35.38 & 69.62 \\
        GLM-4-Voice~\cite{zeng2024glm} & 3.97 & 3.42 & 25.92 & 3.18 & 52.41 & 74.51 \\
        Kimi-Audio~\cite{ding2025kimi} & 4.46 & \textbf{3.97} & \textbf{61.10} & \textbf{4.20} & 59.22 & \textbf{76.96} \\
        LLaMA-Omni2{\tablefootnote{We compare with LLaMA-Omni2-7B-Bilingual for its comparable parameter size and bilingual capability.}}~\cite{fang2025llama} & 3.96 & 3.46 & 17.36 & 3.07 & 39.46 & 63.79 \\
        \texttt{OpenS2S} & 4.09 & 3.65 & 42.89 & 3.66 & 46.90 & 67.68 \\
        \rowcolor{yellow} \texttt{OpenS2S\_V1.5} & \textbf{4.51} & 3.88 & 41.75 & 3.78 & \textbf{59.32} & 72.83 \\
        \bottomrule[1.2pt]
    \end{tabular}
\end{table}

The results in Table~\ref{tab_speech2text} show that \texttt{OpenS2S} demonstrates competitive performance across the four subsets of VoiceBench. Its scores rank second only to Kimi-Audio, which is trained on substantially more data{\footnote{For example, Kimi-Audio employs more 13 million hours of audio data for pre-training.}}, and outperform all other models. These findings indicate that \texttt{OpenS2S} possesses strong capabilities in spoken dialogue and can effectively handle user voice command inputs.

In addition, the results on the URO-Bench subsets show that \texttt{OpenS2S} achieves scores close to those of state-of-the-art models in empathy evaluation, despite being trained on significantly less data. This not only confirms the solid empathetic interaction capabilities of \texttt{OpenS2S}, but also highlights the high quality of the data generated by the proposed empathetic speech dialogue data generation method.

\textcolor{red}{\textbf{UPDATE (October 2025)}} We test the newly released \texttt{OpenS2S\_V1.5}, which is based on \texttt{OpenS2S} and continues the original method for further fine-tuning. This model only extend the data scale during the Empathetic Speech Instruction Tuning phase. The results show that \texttt{OpenS2S\_V1.5} shows significant improvements across most metrics, with two metrics surpassing all baseline models. The results after scaling the training data further validate the effectiveness of the data generation method we proposed.

\subsection{Speech-to-Speech Chat}

Finally, we assess the end-to-end speech conversation capabilities of \texttt{OpenS2S} based on qualitative analysis. Visit~\url{https://casia-lm.github.io/OpenS2S} for demos.

\section{Related Work}

\subsection{Speech Language Models}

With the rapid advancement of large language models (LLMs), there is growing interest in extending their capabilities to spoken language, giving rise to Speech Language Models (SpeechLMs) that can understand and/or generate speech~\cite{cui2024recent, ji2024wavchat}. One approach directly adapts LLMs for \emph{end-to-end speech modeling} by converting speech into discrete tokens and expanding the vocabulary, as seen in SpeechGPT~\cite{zhang2023speechgpt}, AudioPaLM~\cite{rubenstein2023audiopalm}, and TWIST~\cite{hassid2023textually}. Recent models like Spirit-LM~\cite{nguyen2025spirit} and GLM-4-Voice~\cite{zeng2024glm} leverage interleaved speech-text training, while others such as Moshi~\cite{defossez2024moshi} and LSLM~\cite{ma2025language} enable spoken dialogue. In contrast, modular SpeechLMs \emph{connect LLMs with external speech modules}. Early works~\cite{shu2023llasm,wang2023blsp,wang2024blspkd,chu2023qwen,nachmani2023spoken,tang2023salmonn,chu2024qwen2,hu2024wavllm} focused on speech understanding by connecting pretrained speech encoders to LLMs, but did not support speech generation. In contrast, more recent models such as LLaMA-Omni~\cite{fang2024llama,fang2025llama}, Freeze-Omni~\cite{wang2024freeze}, and OpenOmni~\cite{luo2025openomni} overcome this limitation by attaching speech decoders to LLM outputs. Mini-Omni~\cite{xie2024mini} and SLAM-Omni~\cite{chen2024slam} go further with parallel decoding. Minmo~\cite{chen2025minmo} and LLaMA-Omni2~\cite{fang2025llama} incorporates a streaming speech decoder through interleaved text-speech generation.

\subsection{Empathetic Conversations Across Modalities}

Empathetic conversation modeling~\cite{rashkin2018towards,liu2021towards} has been studied across \emph{text-to-text, speech-to-text, and speech-to-speech} settings, aiming to equip LLMs with emotional understanding and supportive responses~\cite{burleson2003emotional}. In text-based interactions, early work focused on architecture modifications~\cite{goel2021emotion}, while recent approaches like SoulChat~\cite{chen2023soulchat} and Chain of Empathy prompting~\cite{lee2023chain} enhance empathy through fine-tuning or step-by-step reasoning without extra data. For speech-to-text interaction, E-chat~\cite{xue2024chat} introduced an emotion-aware speech instruction dataset to enhance LLMs' understanding of emotional speech. BLSP-Emo~\cite{wang2024blsp} proposed an end-to-end model that aligns speech semantics and emotions through two-stage pretraining using ASR and SER datasets. Moving toward speech-to-speech empathy, Spoken-GPT~\cite{zhang2024speechgpt} adopts a cascaded framework that listens and responds with expressive, emotionally attuned speech, paving the way for fully empathetic voice agents. 
Advanced commercial models such as GPT-4o~\cite{hurst2024gpt}, Doubao, Kimi-Audio~\cite{ding2025kimi}, and Step-Audio~\cite{huang2025step} push the boundaries of empathetic interaction by incorporating paralinguistic cues to better perceive and respond to users' emotional states. These models integrate speech understanding and generation in real time, enabling more natural and emotionally aware human-computer interactions. However, our \texttt{OpenS2S} is the first to release all the resources including model weights, training data and training codes, in order to boost the research in the community.

\section{Conclusion}

This report presents \texttt{OpenS2S}, a fully open-source, end-to-end LSLM specifically designed for empathetic speech interactions. 
\texttt{OpenS2S} distinguishes itself with an efficient streaming interleaved decoding architecture, enabling low-latency response generation, and an innovative automated data construction pipeline. 
This pipeline cost-effectively synthesizes diverse, high-quality empathetic speech dialogues by leveraging large language models and controllable text-to-speech systems. 
As a result, \texttt{OpenS2S} achieves competitive performance in empathetic interactions while requiring substantially less data and computational resources compared to current resource-intensive pre-training methods. 
We release the complete \texttt{OpenS2S} framework, including the dataset, model weights, and training codes, to empower the broader research community and accelerate innovation in empathetic speech systems.

\clearpage
\bibliography{reference}
\bibliographystyle{unsrtnat}


\appendix


\end{document}